\documentclass[bst/sn-mathphys-num]{sn-jnl}

\usepackage{graphicx}%
\usepackage{multirow}%
\usepackage{amsmath,amssymb,amsfonts}%
\usepackage{amsthm}%
\usepackage{mathrsfs}%
\usepackage[title]{appendix}%
\usepackage{xcolor}%
\usepackage{soul}
\usepackage{textcomp}%
\usepackage{manyfoot}%
\usepackage{booktabs}%
\usepackage{algorithm}%
\usepackage{algorithmicx}%
\usepackage{algpseudocode}%
\usepackage{listings}%

\theoremstyle{thmstyleone}%
\theoremstyle{thmstyletwo}%

\theoremstyle{thmstylethree}%

\begin{document}

\title[Article Title]{Cost-Efficient Prompt Engineering for
Unsupervised Entity Resolution}


\author[1]{\fnm{Navapat} \sur{Nananukul}}\email{nananuku@isi.edu}

\author[1]{\fnm{Khanin} \sur{Sisaengsuwanchai}}\email{sisaengs@isi.edu}

\author[1]{\fnm{Mayank} \sur{Kejriwal}}\email{kejriwal@isi.edu}

\affil[1]{ \orgname{University of Southern California, Information Sciences Institute}, \city{Los Angeles}, \state{CA}, \country{United States of America}}



\abstract{Entity Resolution (ER) is the problem of semi-automatically determining when two entities refer to the same \emph{underlying} entity, with applications ranging from healthcare to e-commerce. Traditional ER solutions required considerable manual expertise, including domain-specific feature engineering, as well as identification and curation of training data. Recently released large language models (LLMs) provide an opportunity to make ER more seamless and domain-independent. However, it is also well known that LLMs can pose risks, and that the quality of their outputs can depend on how prompts are engineered. Unfortunately, a systematic experimental study on the effects of different prompting methods for addressing unsupervised ER, using LLMs like ChatGPT, has been lacking thus
far. This paper aims to address this gap by conducting such a study. We consider some relatively simple and cost-efficient ER prompt engineering methods and apply them to ER on two real-world datasets widely used in the community. We use an extensive set of experimental results to show that an LLM like GPT3.5 is viable for high-performing unsupervised ER, and interestingly, that more complicated and detailed (and hence, expensive) prompting methods do not necessarily outperform simpler approaches. We provide brief discussions on  qualitative and error analysis, including a study of the inter-consistency of different prompting methods to determine whether they yield stable outputs. Finally, we consider some limitations of LLMs when applied to ER.}

\keywords{large language models, prompt engineering, unsupervised entity resolution, inter-consistency of prompting}



\maketitle

\section{Introduction}

\begin{figure*}[ht]
    \centering\footnotesize
    \includegraphics[width=\textwidth]{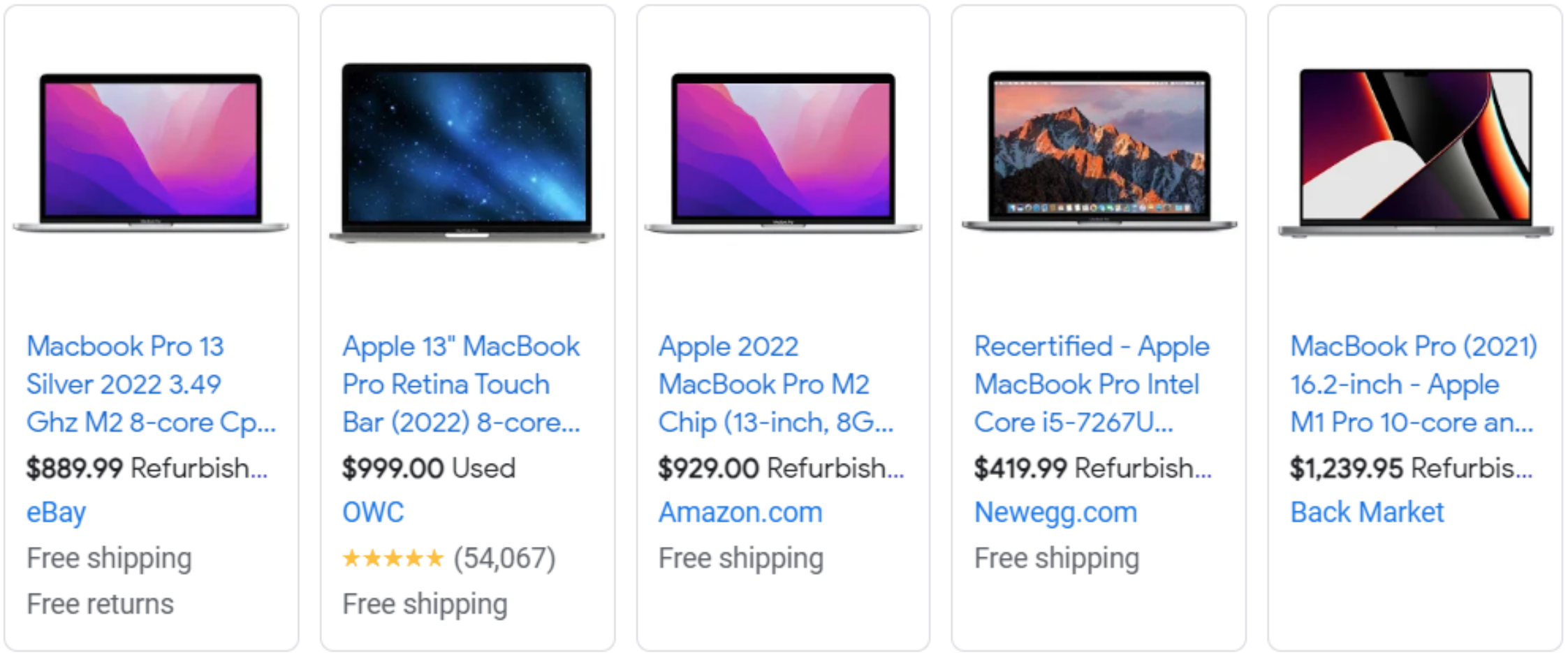}
    \caption{An illustrative example of products from Google Product Search offering the same product from different sources} 
    \label{fig:product-example}
\end{figure*}

Entity Resolution (ER) is (at least) a 50 year-old problem that has been studied in many real-world domains \cite{ERsurvey, kejriwal2023named, ER1, ER2, ER3}. ER can be defined as the algorithmic problem of determining when two or more entities refer to the same underlying entity. Ironically, ER itself has been studied under many names, including entity matching, instance matching, deduplication, and record linkage, to just name a few \cite{elmagarmid2006duplicate}. In real-world domains like healthcare and e-commerce \cite{ERsurvey}, ER can be a complex problem (sometimes, even for human beings). Figure \ref{fig:product-example} provides a representative example in e-commerce. 

Many promising solutions to ER have been proposed over the decades, and significant progress has been achieved. Nevertheless, we are far from solving the problem at human performance levels, and errors can be costly. While initially (and in some contexts still), rule-based and manually engineered solutions were prevalent \cite{papadakis2021four,volz2009silk}, machine learning methods became increasingly popular as ER solutions, as with many other problems amenable to learning from data, starting from the early-mid 2000s \cite{talburt2011entity, kejriwal2016populating}. Over the last decade, deep learning solutions have also been applied to ER \cite{ebraheem2017deeper,DLER1,DLER2,DLER3,DLER4}, and even more recently, transformer-based models such as BERT \cite{ditto}. While performance has steadily improved, general performance on difficult datasets remains subpar.

At the same time, because of the domain-specific nature of ER (e.g., ER in e-commerce \cite{gheini2019unsupervised} can be significantly different than in a healthcare setting \cite{ERhealthcare}, where precision requirements might be stronger), manual engineering continues to play an important role, requiring considerable effort in setting up and executing an ER architecture. One issue is that a typical ER workflow comprises two steps: \emph{blocking} and \emph{similarity} \cite{li2020survey,balaji2016ensemble}. Blocking is necessary to mitigate the quadratic complexity of ER. This problem arises because, given $n$ entities or `records', and a matching function $f$ that indicates whether a pair of records matches or not, a naive ER solution would need $O(n^2)$ comparisons, which is not realistic \cite{blocking1,blocking2}. A blocking function clusters \textit{aproximately similar} records together in \textit{sub-quadratic} runtime, at least in practice. Only records within these clusters, called blocks, are then compared with one another using a similarity function that is typically trained using machine or deep learning. 

In recent years, there has been enormous progress on learning-based and even unsupervised blocking \cite{blocking3}, but it is less clear how to apply LLMs to blocking. In this study, we focus solely on similarity, as our intent is to study whether a large language model (LLM) can be used as a good similarity method. However, keeping the expensive nature of ER in mind, one of our goals is to consider a range of reasonable prompting methods (as similarity functions), some of which are more expensive but may also provide higher performance. For instance, we consider whether adding a persona to the prompt, or providing more detailed instructions (or more finely structured inputs), all of which lead to more tokens in the input and hence, higher \textit{per-prompt} cost, end up yielding proportionately better performance. 

Somewhat more broadly, we also seek to explore whether an established, widely-in-use LLM like ChatGPT offers a promising unsupervised technique of telling when two records are the same or not, at least in a significant application area such as e-commerce. However, clear evidence of this has been lacking in the literature due to the recency of the LLMs. At the same time, because of the growing literature on prompt engineering \cite{prompt1,prompt2,sorensen2022information,wang2023prompt,NEURIPS2022_c4025018}, there is considerable evidence showing that LLMs can be sensitive to the manner in which they are prompted. 

Motivated by both of these observations, we propose an empirical case study that aims to provide systematic guidance on how prompt engineering affects the performance of an LLM like GPT-3.5 on `unsupervised' ER (where no explicit training or fine-tuning is conducted). Specific experimental questions that we consider include: what is the general performance of GPT-3.5 on ER? Is it the case that more expensive and detailed prompt engineering methods necessarily outperform simpler, less expensive and more obvious methods? We investigate these questions using six prompt engineering methods on two real-world e-commerce benchmark datasets that have been extensively used in the literature. At least one of these (based on matching products across Google and Amazon) has been found to be challenging even for state-of-the-art ER systems \cite{peeters2023entity, supervised-contrast, probing-em}.  

The rest of this article is structured as follows. First, we begin with a discussion of related work in Section \ref{sec:rw}, followed by some background and preliminaries on ER in Section \ref{sec:background}. Next, we describe six different prompting `patterns' (or methodologies) for conducting ER in Section \ref{sec:prompt}. These prompting patterns form the basis of the case study. We describe the actual experimental study and results in Section \ref{sec:experiments}, before concluding the article in Section \ref{sec:conclusion}.

\section{Related Work}\label{sec:rw}


\textbf{Entity Resolution.} Entity Resolution (ER) is a fundamental task in data management and quality, aimed at identifying and linking records that refer to the same real-world entity across different data sources, or even within a single source. Typically, ER methods for structured data \cite{elmagarmid2006duplicate} include at least a \textit{blocking} step and a \textit{matching} step. This paper focuses on the latter, where the matching step partitions
a candidate set of pairs (obtained through blocking) into matching pairs (duplicates) and non-matching pairs (non-duplicates). Traditional methods for performing ER tend to be grouped into two main categories: \textit{machine learning-based} and \textit{rule-based}. 

Previous research has highlighted machine learning-based methods for ER, utilizing classifiers trained on labeled datasets to identify duplicates. Commonly used algorithms in the past have included \textit{Support Vector Machines (SVM), Naive Bayes,} and \textit{Decision Trees} \cite{christen2008febrl,bilenko2003adaptive,cohen2002learning,sarawagi2002interactive}
These learning-based methods first extract features from each record pair in the candidate set using the full set of attributes in the dataset's schema and uses the extracted features to train a binary classifier. One important step required for the learning-based methods is selection of appropriate classification features for the model (feature engineering). Feature engineering is inherently labor-intensive and expertise-driven, often becoming a bottleneck in this ER process. More recently, deep learning ER methods have proposed to eliminate this step through automatic (neural network-based) representation learning \cite{ebraheem2017deeper,gottapu2016entity}.

Rule-based methods \cite{fan2009reasoning,shen2005constraint,singla2006entity} compute the similarity between corresponding attribute values using similarity metrics \cite{cohen2003comparison}. The final ER decisions are based on pre-defined rules and thresholds. Since rule-based methods rely on the establishment of explicit rules and thresholds that determine how entities are matched, feature engineering is not required; however, the formulation of rules can still be a cumbersome process and may not be data-driven or optimal. Hence, the efficacy of rule-based methods significantly depends on the expertise of domain specialists who devise and fine-tune the specific rules and thresholds. The involvement of these experts is critical to ensuring that the rules accurately reflect the nuances and complexities of the specific domain of the datasets.

\textbf{Large Language Models (LLMs).} Recent years have seen significant progress on generative AI models, of which the large language models like ChatGPT mark an important milestone. Multiple LLMs have been developed and released recently. Notable examples include OpenAI's ChatGPT \cite{openai2024gpt4}, Meta's LLaMA \cite{touvron2023llama}, and Google's Gemini \cite{geminiteam2023gemini}. LLMs have rapidly become popular for a range of problems involving both text (and more recently, multi-modal inputs) and have proven adept at (i) processing natural language context, (ii) generating human-like text, and (iii) using the trained knowledge-intensive corpus  (on which they have been pretrained) to engage in `in-context' learning and respond to a variety of prompts. It is this last ability of LLMs that we study in this paper specifically in the context of nearly-unsupervised ER. Applications of LLMs span across multiple domains related to natural language processing, e.g., search \cite{arcila2023platform}, customer support \cite{spatharioti2023comparing}, and translation \cite{karpinska2023large}, and continue to grow. Even in more domain-specific or specialized scenarios, such as healthcare \cite{wang2023prompt,mesko2023prompt}, academic writing \cite{giray2023prompt}, and visual generation \cite{shtedritski2023does}, LLMs have tended to perform competitively, and have augmented human abilities in impressive ways. 

The use of LLMs for ER is a new area of study. An obvious approach is to use LLMs to act as a similarity measure since they have pre-trained knowledge about real-world entities. Following this approach, LLMs provide a possibility to mitigate some of the inherent challenges of traditional methods, such as feature engineering and rule-construction by domain experts. While LLMs do require `prompt engineering', they are usually robust to prompts in many task-areas. Nonetheless, the manner in which the LLM is prompted can make a difference, as we also explore in this paper. In related work on using LLMs for ER and other such problems, recent LLM-based research in knowledge graph construction has explored how prompt engineering techniques, such as few-shot prompting \cite{logan2021cutting}, can enhance information extraction using LLMs \cite{schacht2023promptie,polak2023extracting,hu2024improving}. Another paper proposed a knowledge graph construction framework by performing entity extraction and triple generation \cite{bi2024codekgc}. To our knowledge, there is a gap in research regarding the use of LLMs as a similarity function for entity resolution, as well as prompt engineering LLMs for ER. Recognizing this gap, we provide a detailed case study in this paper using GPT-3.5 for the matching step. We also report insights into how different prompting patterns affect GPT-3.5's ER performance relative to the cost, and supplement our quantitative results with qualitative analyses using responses obtained from GPT-3.5.

\section{Background: Entity Resolution}\label{sec:background}

This section provides some technical background on ER to lay the groundwork for the remainder of the article. As intuited earlier, the goal of ER is to identify which records (also called ``entity profiles'' or ``mentions'' \cite{papadakis2020three,bhattacharya2006entity}) from one or more data sources refer to the same real-world entity. The challenge arises from the fact that the same real-world entity can be represented in different ways, making it difficult for machines to follow a rigid set of rules for matching entities. In contrast, the problem is easy for most humans in common domains. 

To study the problem algorithmically, we begin by defining an \textit{entity profile} $e_{id}$ as a pair $(id, A_{id})$, where $id$ is a unique identifier for the profile, and $A_{id}$ is a \textit{dictionary} of attributes describing each profile i.e., $A_{id}$ may be expressed as a set of key-value pairs: $(attribute\_name, attribute\_value)$. We denote a set $E$ of entity profiles as an \textit{entity collection}. We provide examples of entity profile representations of the first two laptops in Figure \ref{fig:product-example}. 

Given an entity collection $E$, the task of ER is to partition the entity profiles in $E$ into clusters, such that each cluster represents a unique entity \cite{getoor2012entity,christendata}. This general problem affords many variants, the most common one of which is to find \textit{pairs} of entity profiles (rather than clusters), such that each pair represents a `match' or a `duplicate'. One reason for interpreting ER as a pairwise problem is that, given matches, often with accompanying similarity scores, graph clustering algorithms can be used to combine matches into clusters. In many applications, matching pairs are enough and more general clusters are not needed. Hence, most ER benchmarks are set up to evaluate the pairwise version of the problem. We assume the same here as well. 

Formally, we define the task of ER as discovering all duplicates ($e_{i}$, $e_{j}$), with $e_{i}, e_{j} \in E$. Such (predicted) duplicates are then evaluated with respect to a \textit{ground-truth} of (typically, manually labeled) duplicates, similar to other problems in machine learning. In its supervised variant, some fraction of these manually labeled pairs are provided to the algorithm as input, which then has to learn from them and predict the remaining duplicates. Any pair not labeled as a duplicate is automatically considered as a non-duplicate. One other point to note is that, although we assumed a single entity collection $E$, in practice, it is not uncommon to assume \textit{two} entity collections $E_1$ and $E_2$ that are individually de-duplicated, but that have matching entities between them. We see this situation reflected in Figure \ref{fig:product-example}: the laptop from eBay would need to be matched to the laptop from OWC, the laptop from Amazon.com, and so on. If two entity collections (e.g., eBay and Amazon) are assumed, as in this article, then the problem of ER can be defined analogously as discovering all duplicates ($e_{i}$, $e_{j}$) between $E_1$ and $E_2$ ($e_{i} \in E_1$, $e_{j} \in E_2$).

The performance of an ER system is measured both by its effectiveness and efficiency. \textit{Effectiveness} refers to the number of actual duplicates that the ER algorithm is able to find (among all the duplicates in the ground-truth), while minimizing the number of false positives (predicted duplicates that are actually non-duplicates). It is measured using metrics like precision, recall, and F-measure, and that we briefly describe in Section \ref{sec:experiments}. \textit{Efficiency} refers to the system's computational cost, which is quadratic in the worst case ($O(|E_1||E_2|)$), even if the matching or `similarity function' is known beforehand. If the function ran in constant time, a brute-force approach would still have to perform \textit{all} pairwise comparisons between the collections, which quickly runs into the millions even for a few thousand entity profiles in each collection. 

To mitigate this brute-force quadratic complexity, the ER community has developed a family of approaches called \textit{blocking} that clusters approximately similar entities into blocks, typically using indexing-like algorithms that run in near-linear time. Even simple blocking can help eliminate a large fraction of unnecessary comparisons. For example, given entity profiles representing people, some of which are duplicates, a blocking algorithm might use the birthyear and the first two digits of the person's address-zipcode to only compare pairs of people with both of these `blocking keys' in common. Many sophisticated blocking algorithms now exist; we refer the interested reader to \cite{papadakis2020blocking,li2020survey} for comprehensive surveys and approaches.  


The final ER result is determined by using the similarity function on pairs of entity profiles that share blocks (indexed by blocking keys, such as in the example above). Over the decades, many different types of similarity functions have been proposed, ranging from rule-based approaches to string similarity functions (e.g., edit distance) and more recently, machine learning and deep learning (as discussed earlier in the related work) \cite{mudgal2018deep,nie2019deep,zhao2019auto}. However, a systematic evaluation of LLMs as similarity functions, especially when prompted in different but reasonable ways, has been lacking. Next, we describe six such prompting methodologies that can be used as LLM-based in-context similarity functions for low-supervision ER.   




\section{Prompt Engineering Methods for ER}\label{sec:prompt}



This section discusses the the prompt engineering methods we used to study GPT-3.5's ER performance. We begin by describing the underlying ER workflow (when an LLM like GPT-3.5 is used as the similarity function), followed by six specific prompting `patterns' underlying our experiments.

\begin{figure*}[t]
    \centering\footnotesize
    \includegraphics[width=\textwidth]{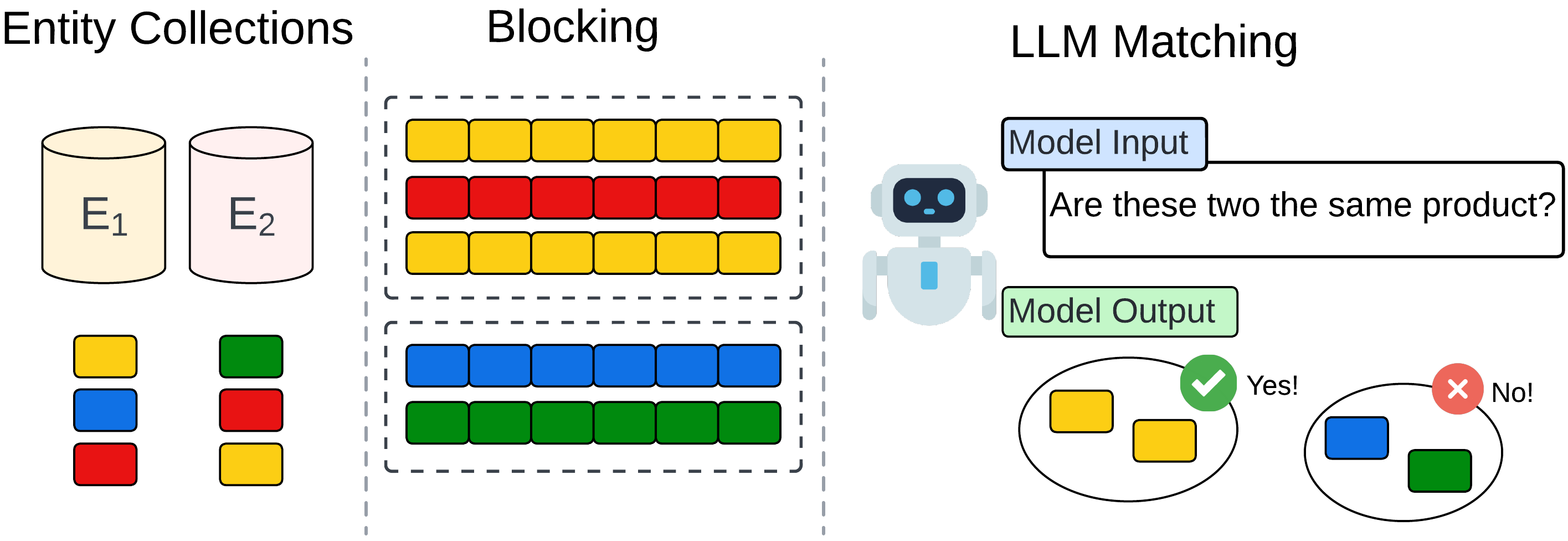}
    \caption{An illustrative example of a typical ER workflow, but with an LLM used as the similarity (or `matching') function. Each entity collection represents a structured ER dataset, with individual entities represented using colored boxes. As explained in Section \ref{sec:background}, blocking is first applied to cluster approximately similar entities into blocks, in order to mitigate the quadratic complexity of comparing all pairs of entities. Only entities sharing a block are paired and presented to the LLM for making a final decision on whether they match (yes) or not (no). } 
    \label{fig:er-llm}
\end{figure*}

Figure \ref{fig:er-llm} illustrates an ER workflow where an LLM serves as the similarity function. While the LLM is technically a `black box' in that it is infeasible to re-train it from scratch, or (without training data) fine-tune it, it can be controlled through prompting. One example is \textit{few-shot prompting}, a technique that integrates a small set of question-and-answer examples, referred to as \textit{demonstrations}, into the prompts \cite{logan2021cutting}. Few-shot prompting enables in-context learning by guiding LLMs to respond based on the patterns and logic observed in the provided demonstrations. In more advanced techniques, such as \textit{Chain-of-Thought (CoT) prompting} \cite{wei2022chain,wang2022self}, intermediate reasoning steps are added to question-and-answer demonstrations. This technique directs LLMs to mimic these reasoning steps when generating responses. Prompt engineering is the process of designing, refining, and implementing prompts that guide the LLMs to output better results \cite{ekin2023prompt}. The original approach in the prompt engineering process involves crafting the main components of the prompt, such as prompt instruction, context/persona, and output format. Despite the simplicity of its implementation, effective prompt engineering is a non-trivial process, especially if we take both costs and benefits into account. For instance, it is not always clear that longer prompts yield better performance than short prompts, even though the former is more expensive (involving more tokens) than the latter. 
Prompt engineering has been actively researched in domains ranging from graph analytics \cite{fatemi2023talk} and healthcare \cite{mesko2023prompt} to business process management \cite{busch2023just}, but to our knowledge, remains to be systematically studied in ER.

\begin{figure*}[ht]
    \centering\footnotesize
    \includegraphics[width=0.95\textwidth]{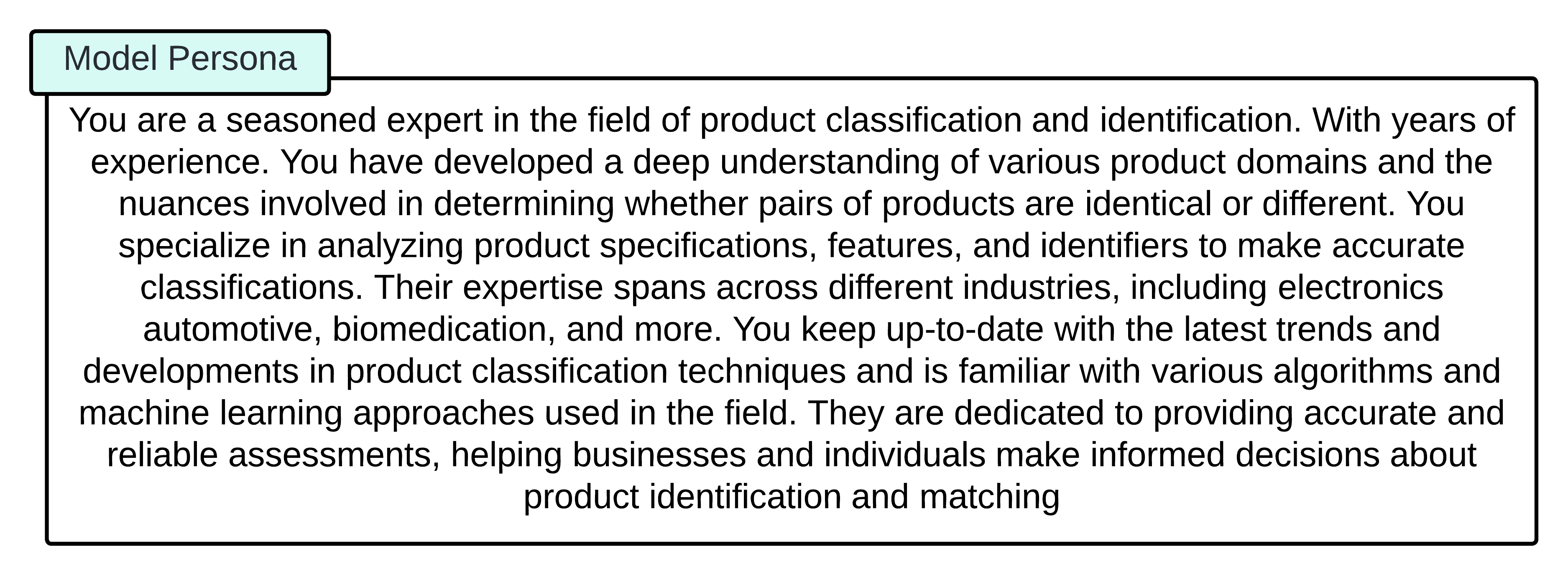}
    \caption{A text description of the \textit{persona} describing the conceptual role of GPT-3.5 when performing ER. Since this study uses a dataset in the product domain, we set up the role of GPT-3.5 as an expert on product classification, with an emphasis on using related knowledge from multiple product domains to resolve the pair.} 
    \label{fig:persona}
\end{figure*}

We construct a prompt template to guide GPT-3.5 in performing ER. In order for GPT-3.5 to produce ER results, the prompt template must include three main components: a \textit{candidate pair} (that we are seeking to determine is a match or non-match), \textit{ER instructions}, and a \textit{result format}. Modification to these components results in different \textit{prompt patterns} that influence the LLM's ER results. To study the influence of prompting, we construct several such prompt patterns by altering the three main components in the prompt template. Besides altering these main components, prompt patterns can also be generated by appending \textit{optional} components that help influence ER results. We experimented with two such optional components in this study: \textit{persona} and \textit{few-shot examples}. The \textit{persona} component makes the LLM's role more specific by defining a character profile that outlines how the LLM should behave. Users can set up the persona as a preset by adding a text description when using GPT-3.5. Figure \ref{fig:persona} illustrates a text explanation of the ER expert persona we used in this experiment. In contrast, as previously discussed, \textit{few-shot examples} guide the LLM to generate responses by learning in-context from the demonstrations embedded within these examples. In ER, we can use sampled pairs with labels from the ground-truth as demonstrations to generate a few-shot prompt pattern.

In the following subsections, we illustrate these six prompt patterns and describe the rationale behind their creation. Throughout this section, we use a candidate pair in Table \ref{tab:product_comparison}  from the Amazon-Google dataset to illustrate the prompt patterns.

\begin{table}[ht]
\centering
\begin{tabular}{|p{0.5in}|p{1.3in}|p{0.8in}|p{2.5in}|}
\hline
\textbf{Datasets} & \textbf{Product Title} & \textbf{Manufacturer} & \textbf{Product Descriptions} \\
\hline
Amazon & Apple final cut studio 2 (mac) & Apple & final cut studio 2 delivers an integrated post-production solution that lets you move effortlessly \ldots \\
\hline
Google & Apple final cut studio 2 production suite for mac av production software & N/A & final cut studio 2 production software suite for mac - final cut pro 6 motion 3 soundtrack pro 2 color compressor 3 dvd studio pro 4 \ldots \\
\hline
\end{tabular}
\caption{Example of a candidate pair from the Amazon-Google dataset featuring the product `Apple final cut studio 2', with key attributes displayed: \textit{product title, manufacturer, }and\textit{ product description}.}
\label{tab:product_comparison}
\end{table}


\subsection{Single-Attribute Prompt Pattern}

\begin{figure*}[ht]
    \centering\footnotesize
    \includegraphics[width=\textwidth]{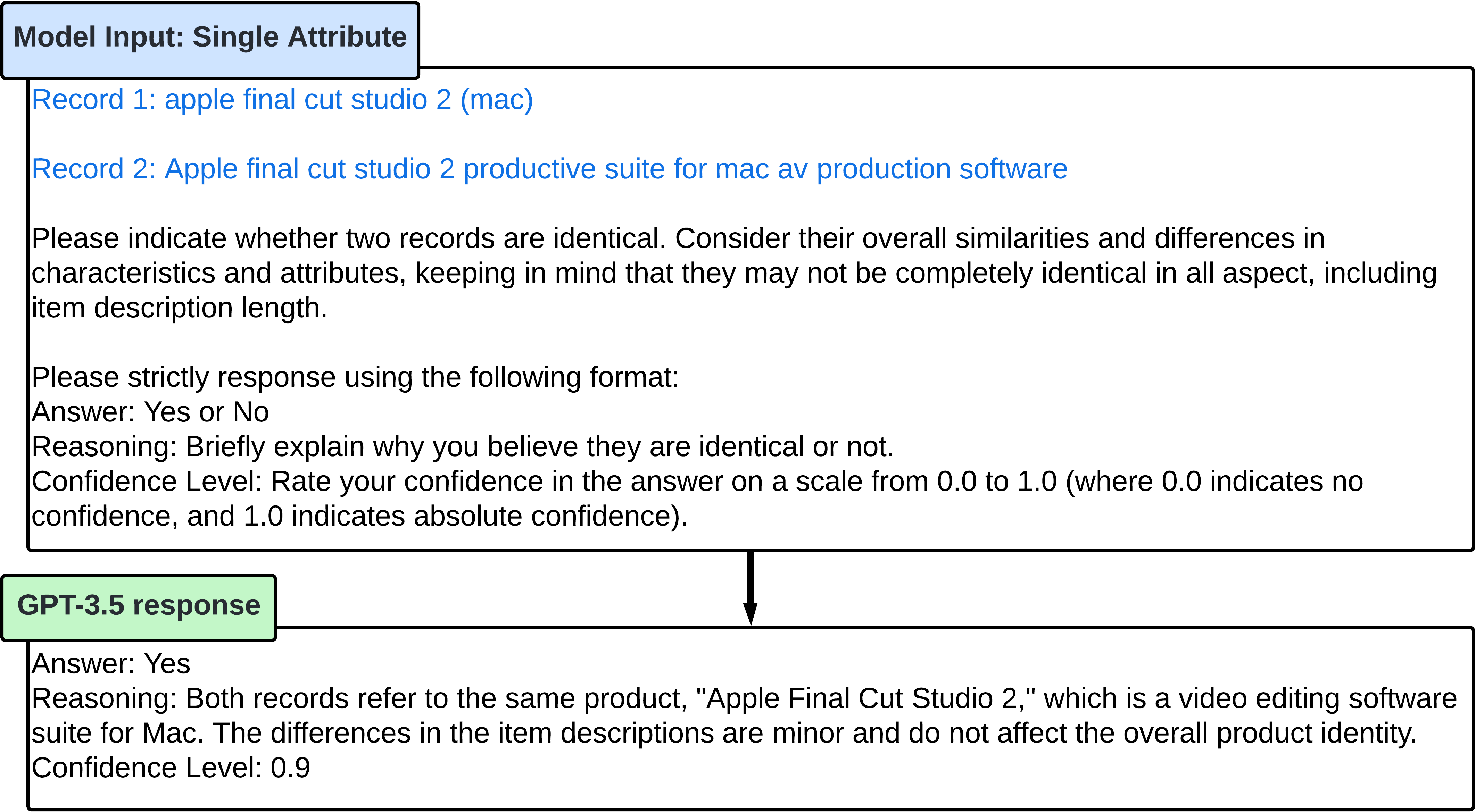}
    \caption{An illustrative example of the \emph{single-attr} prompt pattern. The example includes two records for comparison: both are versions of `Apple Final Cut Studio 2'. The records information only contains product titles (highlighted in blue). The task instruction asks GPT-3.5 to determine if the records refer to identical products, considering their similarities and differences. The GPT-3.5 response shows that the records are identical and explains the reasoning behind the decision, along with an associated confidence level.} 
    \label{fig:single}
\end{figure*}

The \emph{single-attribute} (henceforth, \emph{single-attr}) is the simplest and cheapest prompting pattern out of the six patterns. This pattern assumes that there is a single attribute containing the appropriate information to enable the model to make a good matching decision, and that we (as human prompters) are able to identify this attribute. While the assumption seems restrictive, it is a good baseline, as in many cases, domain experts have an intuition for which attributes contain high `information density'.

In the context of the e-commerce domain, to construct the pattern, we selected the \textit{product title} as the primary attribute for this pattern because of its concise and descriptive nature, which can be useful for the LLM in identifying products within e-commerce benchmark datasets, such as WDC and Amazon-Google Products (subsequently detailed). The advantage of this pattern is its cost efficiency due to significant token reduction by using only one attribute. However, it has the obvious drawback of making mistakes when the information that can differentiate two records happens to be in \textit{other} attributes. The prompt example is shown in Figure \ref{fig:single}, where GPT-3.5 concludes the result based on information from the product title (highlighted in blue).

\begin{figure*}[ht]
    \centering\footnotesize
    \includegraphics[width=\textwidth]{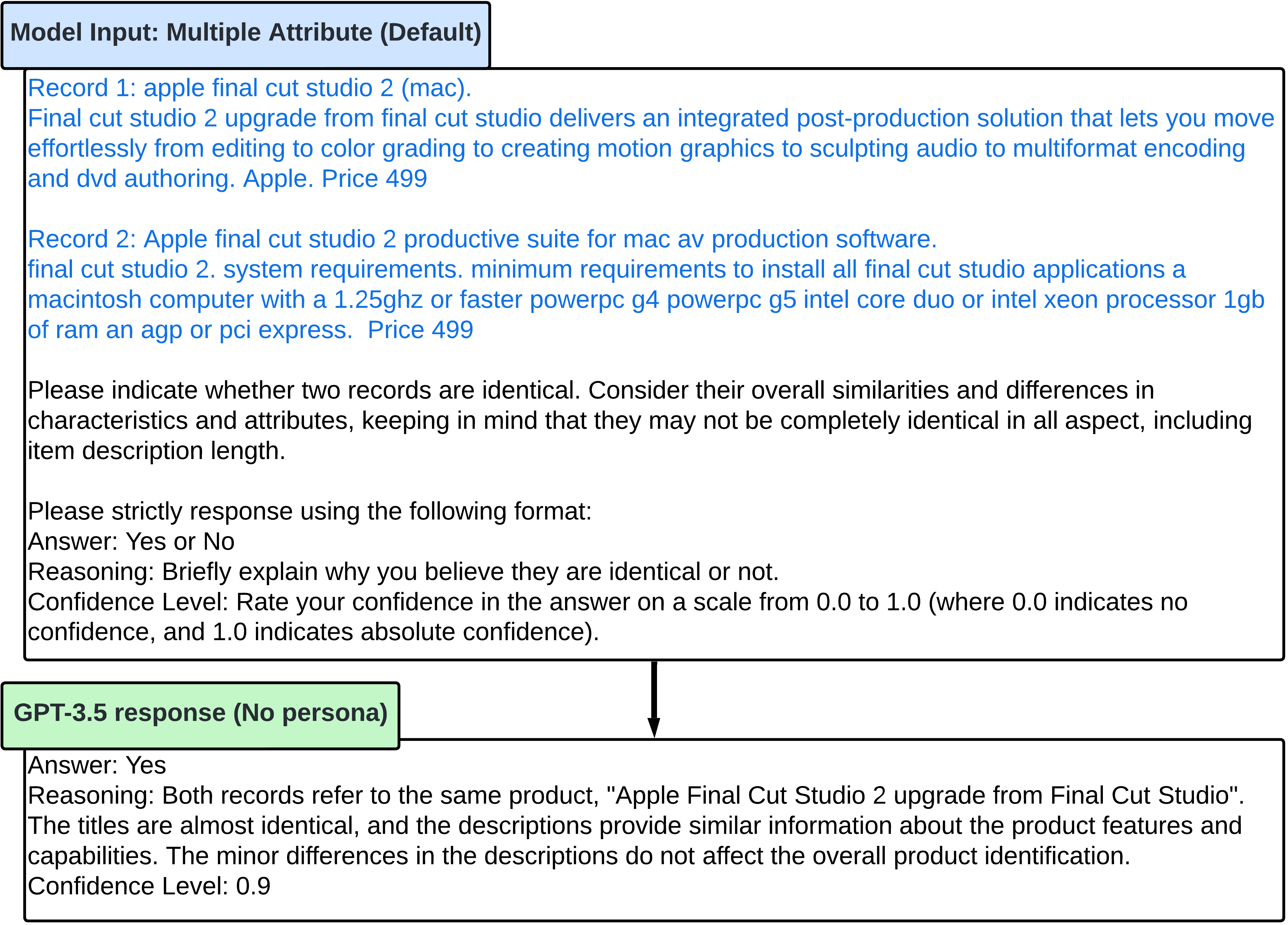}
    \caption{An illustrative example of the \emph{multi-attr} and \emph{no-persona} prompt patterns, comparing records with multiple attributes concatenated (highlighted in blue). As with the \emph{single-attr} pattern, GPT-3.5 is tasked with evaluating whether the detailed product records, now including additional attributes like \textit{manufacturer, description,} and \textit{price}, refer to the same underlying product. The GPT-3.5 \textit{response format} remains the same as the \emph{single-attr} pattern.} 
    \label{fig:multi-default}
\end{figure*}

\subsection{Multi-Attribute Prompt Patterns}

Unlike \emph{single-attr}, which uses one attribute to represent candidate pairs, the subsequent prompt patterns utilize multiple attributes from an ER dataset without any preprocessing. These patterns can be categorized into two methods:  (1) \emph{multiple attributes without using the LLM persona} (henceforth, \emph{no-persona}) and (2) \emph{multiple attributes with persona} (henceforth, \emph{multi-attr}). The distinction between these two patterns is the option to use or omit the ER expert persona. Specifically, the \emph{no-persona} pattern refers to omitting the user-defined textual information that characterizes the LLMs' given `role', as explained earlier in this section. The \emph{no-persona} is the only pattern that omits the persona, while every other method in this study includes a persona in the model. In contrast, \emph{multi-attr} integrates the persona in Figure \ref{fig:persona} with multiple attributes, as demonstrated in Figure \ref{fig:multi-default}, where the candidate pair is presented as a concatenation of all attributes in the dataset.

\begin{figure*}[ht]
    \centering\footnotesize
    \includegraphics[width=\textwidth]{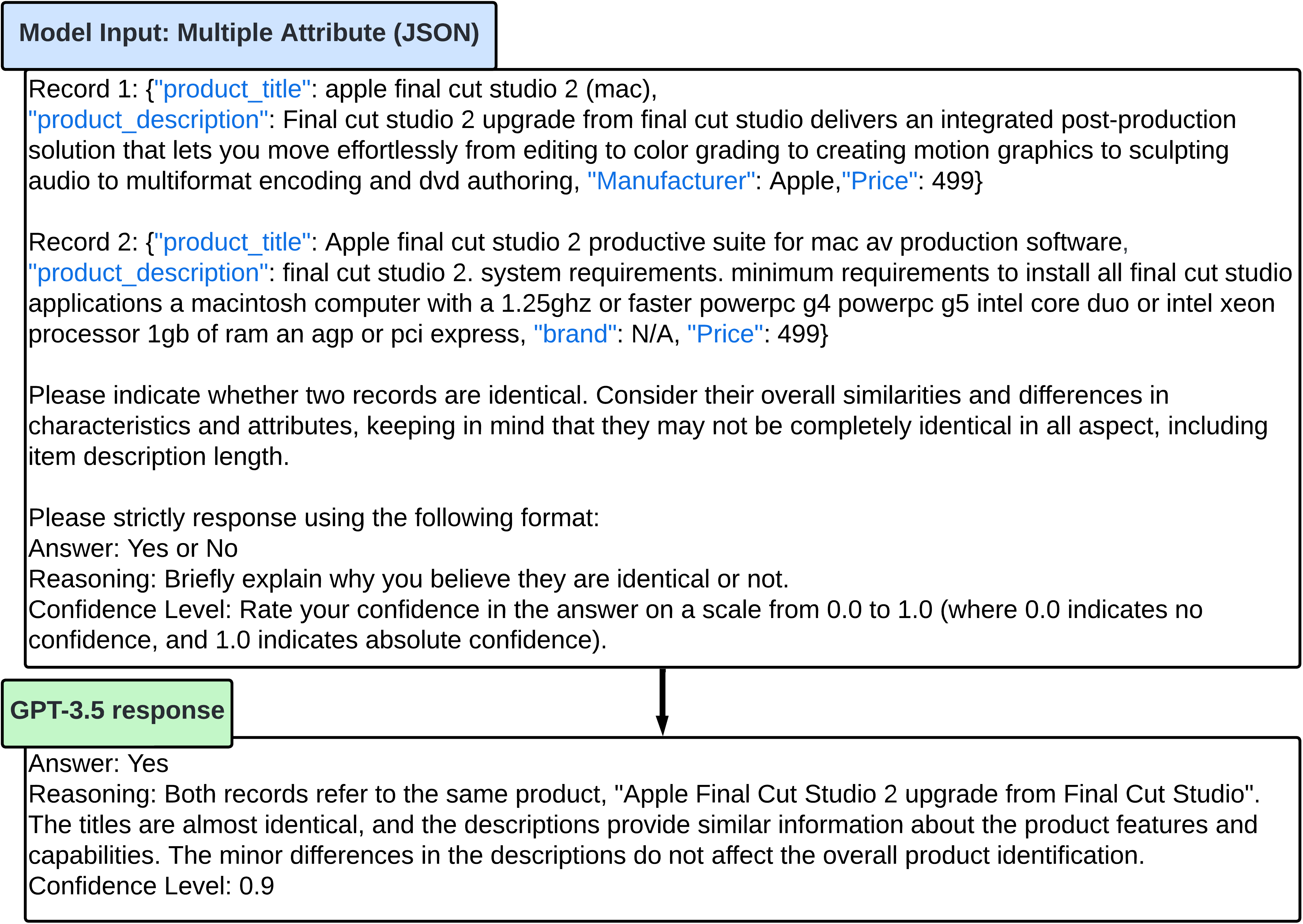}
    \caption{An illustrative example of \emph{multi-json} prompt pattern, showing a product pair in the JavaScript Object Notation (JSON) format. The attributes are organized into a structured format, which includes the \textit{product title, description, manufacturer}, and \textit{price} for each record (JSON keys are highlighted in blue to emphasize the structure). The remaining instructions are the same as those for the \emph{multi-attr} prompt pattern.} 
    \label{fig:multi-json}
\end{figure*}

\subsection{Multi-Attribute with JSON  Prompt Pattern}

One problem with concatenation is that it might make the content ambiguous; hence, we also consider a machine-readable structured version called \emph{multiple attributes with JSON format} (henceforth, \emph{multi-json}) whereby the attributes are represented in a JSON-like format, with the schema information. The format retains the same content as the \emph{multi-attr} method. We hypothesized that a machine-readable format would enhance LLMs' understanding of candidate pairs' attributes, thereby enhancing the LLMs' performance. Figure \ref{fig:multi-json} illustrates an example of JSON-formatted pairs, highlighted in blue, where attribute names serve as keys and are paired with their respective values.

\subsection{Multi-Attribute with Similarity  Prompt Pattern}

\begin{figure*}[ht]
    \centering\footnotesize
    \includegraphics[width=\textwidth]{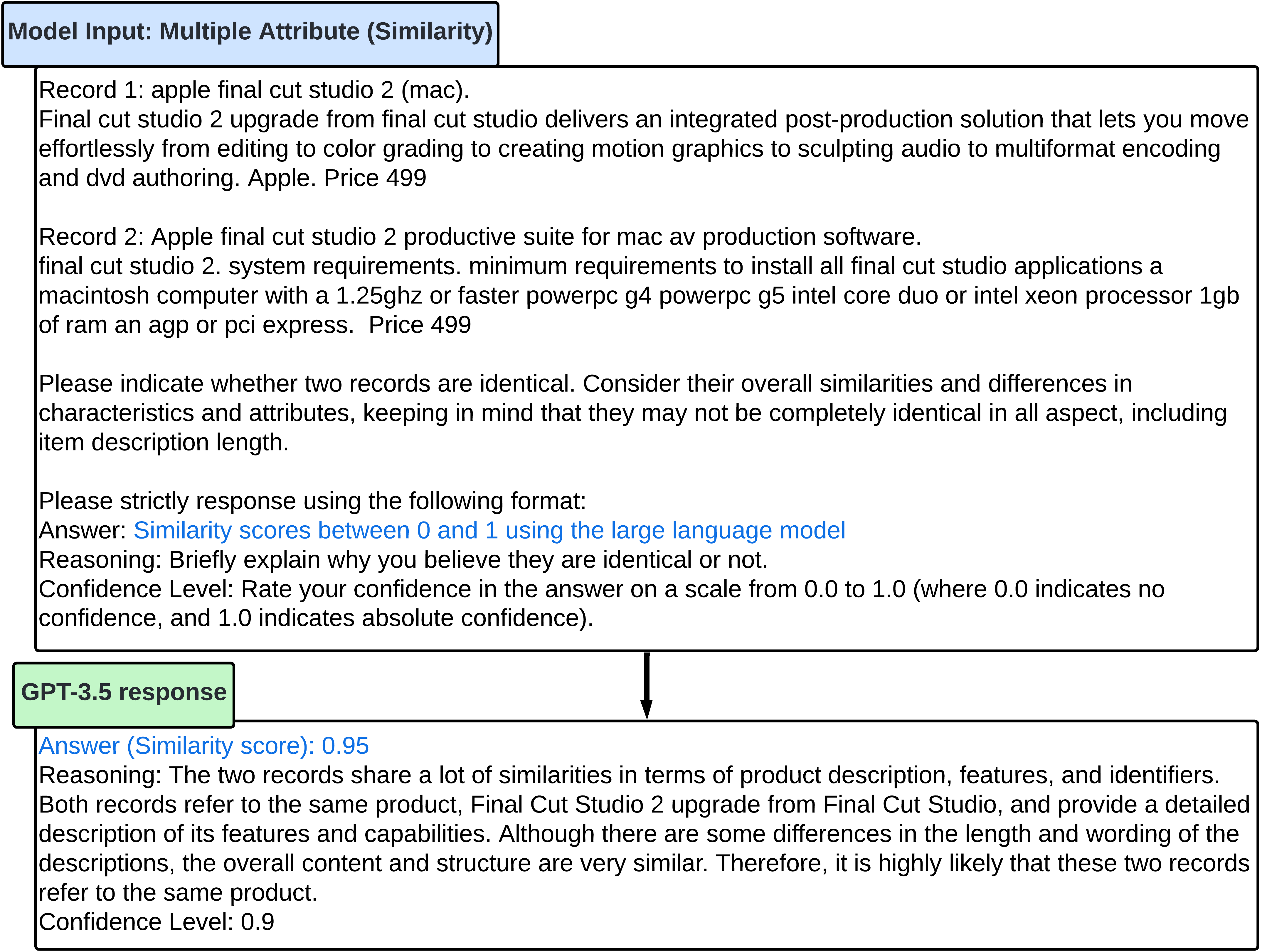}
    \caption{An illustrative example of the \emph{multi-sim} prompt pattern, where the key difference is the instruction for GPT-3.5 to provide a similarity score ranging from 0 to 1 (highlighted in blue). The remaining structure aligns with that of the \emph{multi-attr} pattern.} 
    \label{fig:multi-sim}
\end{figure*}

As discussed thus far, we defined a prompt instruction as directing an LLM to decide whether records are duplicates or non-duplicates. In this prompt pattern (\emph{multiple attributes with similarity}; henceforth, \emph{multi-sim}), we modified the prompt instruction for GPT-3.5 to generate a \textit{similarity score} between two products, instead of giving a definitive ER decision (match or non-match). We did not specify a similarity metric, such as edit distance, in the prompt. Instead, we rely on GPT-3.5's judgment to generate the similarity score on its own. Figure \ref{fig:multi-sim} displays an example of this pattern, highlighting the modified instruction and similarity score in blue. After obtaining similarity scores for each candidate pair, we choose a threshold $\theta$ that yields the best F-1 score for the entire dataset to compute the final ER results.

\subsection{Few-Shot Examples  Prompt Pattern}

\begin{figure*}[ht]
    \centering\footnotesize
    \includegraphics[width=\textwidth]{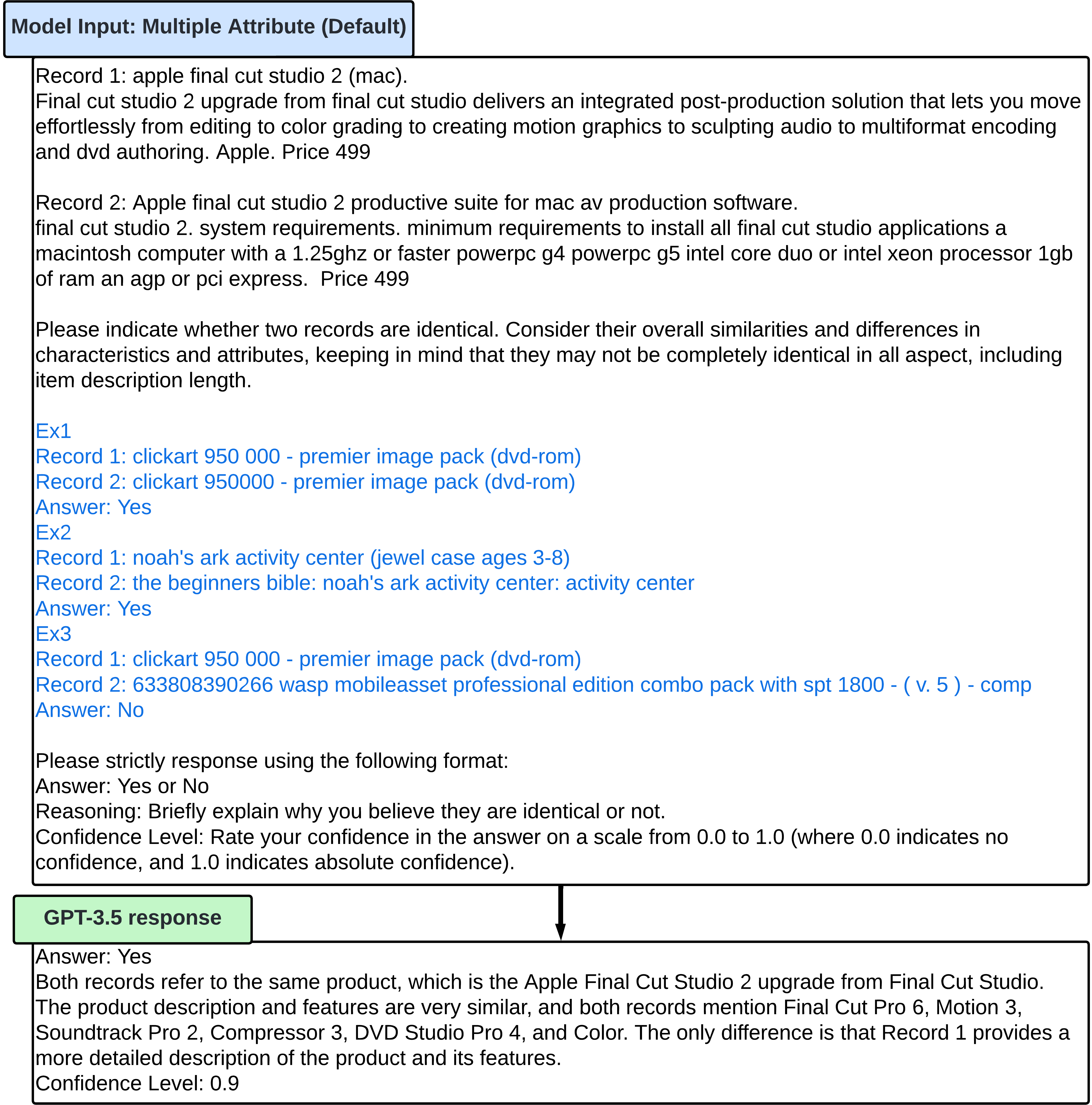}
    \caption{An illustrative example of the \emph{few-shot} prompt pattern, incorporating three examples (two duplicates and one non-duplicate) highlighted in blue. These examples are added to the \emph{multi-attr} prompt pattern, with the rest of the prompt structure remaining unchanged.} 
    \label{fig:few-shot}
\end{figure*}

Finally, we consider a \emph{few-shot} prompting method that is nearly (but not fully) unsupervised. This prompt pattern presents the model with three randomly selected pair examples from the training partition of the corresponding benchmark dataset, which is otherwise unused for the other methods. Figure \ref{fig:few-shot} illustrates few-shot examples, highlighted in blue, where we concatenated two duplicate demonstrations and one non-duplicate demonstration into the original \emph{multi-attr} prompt pattern. As stated earlier, few-shot prompting enables in-context learning to guide LLMs' responses, with the intuition that the model can make a more informed decision if it gets concrete examples of what constitutes duplicate and non-duplicate pairs.

\section{Experimental Study}\label{sec:experiments}

E-commerce is a challenging domain because entities within it are both attribute-rich and can have small differences (such as a 13-inch versus a 15-inch laptop) that make two products different, even though other attributes may be the same. This allows us to check both the robustness and generalization of LLMs like GPT-3.5, which are sometimes known to hallucinate \cite{10319443,chen2023hallucination}, or otherwise focus too much on language compared to the knowledge embedded in the language.  Our experimental design is inspired by previous research \cite{gheini2019unsupervised,vandic2020scalable}, where the ER problem (called \textit{product ER}) takes entity collections of real-world product-profiles as input, and needs to find pairs of profiles referring to the same underlying product. In the datasets described below, thousands of merchants offer a variety of products using different names and descriptions. Products from multiple platforms can have heterogeneous descriptions, including missing values. 

Earlier, we illustrated an example in Figure \ref{fig:product-example} for the Google Product keyword search \textit{Macbook pro 2022}. The search results showed differences in product names and descriptions across different merchants. Not only do different merchants use heterogeneous names and descriptions for describing the same product but the order of details and the way that the text is written is also different. This real-world example underscores the difficulty of product ER. We also considered product ER to be a good application area because studies by other authors \cite{roumeliotis2024llms,farfade2024scaling} have demonstrated LLMs' general capability to reason over e-commerce products in other applications. However, in product ER, LLMs' abilities have not been measured or commented upon in depth.

We use two publicly available and real-world ER benchmarks in the e-commerce domain for this study:

\begin{enumerate}
  \item \textbf{Web Data Commons (WDC)} \cite{primpeli2019wdc}: The full version of this dataset contains 26 million products and descriptions from different e-commerce websites scraped from the Web as part of the Web Data Commons (Common Crawl). The products are categorized based on computers, cameras, shoes, and watches. We selected products of the \emph{computer} type as the test dataset. This smaller curated dataset consists of seven attributes and comprises 1,100 pairs, of which 300 are duplicates and the rest are non-duplicates. 
  \item \textbf{Amazon-Google Products (AG)} \cite{amazon-google-source}: This benchmark contains products based on software and computer hardware from Amazon and Google. The ER linkage must happen between the platforms. The dataset is larger and more challenging, consisting of 11,460 pairs, of which 1,166 are positives. There are only three attributes, one of which is a detailed `description' attribute that is text-heavy. 
\end{enumerate}


For quantifying ER performance, we use standard metrics such as precision, recall, and F-Measure (FM), also called the F1-Score. The precision is the ratio of true positives returned by the ER model to all positives predicted by the model, while the recall is the ratio of true positives returned by the ER model to all positives in the ground-truth. The F-Measure is the harmonic mean of the two quantities, and represents a single-point estimate of a method's ER performance. For the sake of completeness, we also provide formulae for these below:

\begin{equation}
Precision = \frac{|TP|}{|TP|+|FP|}
\end{equation}

\begin{equation}
Recall = \frac{|TP|}{|TP|+|FN|}
\end{equation}

\begin{equation}
F-Measure = \frac{2 \times Precision \times Recall}{Precision + Recall}
\end{equation}
\\
Here, TP, FP, and FN stand for the sets of true positives, false positives, and false negatives returned by the model, respectively. Cost estimates for a complete experiment (i.e., per-benchmark and per-prompting method) are directly obtained from OpenAI's invoices and are also reported. 







\subsection{Results and Discussion} 

\begin{table*}
\begin{center}
\begin{tabular}{|p{0.66in}|p{0.60in}|p{0.39in}|p{0.22in}|p{0.29in}|p{0.60in}|p{0.39in}|p{0.22in}|p{0.29in}|}
    \hline
    \textbf{Prompt Pattern} & \multicolumn{4}{c|}{\textbf{WDC}} & \multicolumn{4}{c|}{\textbf{Amazon-Google}} \\
    \hline
    & \textbf{Precision} & \textbf{Recall} & \textbf{FM} & \textbf{Cost} & \textbf{Precision} & \textbf{Recall} & \textbf{FM} & \textbf{Cost} \\
    \hline
    multi-attr&0.92& 0.90 & 0.91 & \$0.93 & 0.97 & 0.79 & 0.87 & \$3.04 \\
    \hline
    single-attr&0.91& 0.94 & 0.93 & \$0.59 & 0.96 & 0.70 & 0.81 & \$2.19 \\
    \hline
    multi-json& 0.96 & 0.70 & 0.81 & \$0.99 & 0.98 & 0.53 & 0.69 & \$3.23 \\
    \hline
    few-shot& 0.94 & 0.87 & 0.90 & \$1.36 & 0.97 & 0.79 & 0.87 & \$3.75 \\
    \hline
    multi-sim& 0.78 & 0.66 & 0.71 & \$0.95 & 0.93 & 0.97 & 0.95 & \$3.11 \\
    \hline
    no-persona& 0.97 & 0.85 & 0.91 & \$0.68 & 0.97 & 0.56 & 0.71 & \$2.01 \\
    \hline
\end{tabular}
\caption{Evaluation results and cost summary for the six different prompt engineering methods (Section 2) for both benchmarks (WDC and Amazon-Google).}
\label{tab:prompt_eng_result}
\end{center}
\end{table*}

Table \ref{tab:prompt_eng_result} shows the result of each prompt pattern across both benchmarks. Overall, the methods generally perform well, with most methods achieving more than 80\% F-Measure even on the (relatively difficult) Amazon-Google benchmark. As expected, there is a significant \emph{relative} difference in costs (in percentage terms) due to the considerably lower number of per-prompt tokens used by \emph{no-persona} and \emph{single-attr} compared to (for example) \emph{multi-attr} and \emph{few-shot}. Below, we discuss some specific findings derived from the experimental results. 

\textbf{LLMs' performance declined when using JSON-formatted prompts: } Table \ref{tab:prompt_eng_result} shows that the \emph{multi-json} pattern under-performed compared to both \emph{multi-attr} and \emph{multi-sim}, despite containing identical pair information. For example, altering the format from \emph{multi-attr} to \emph{multi-json} led to a drop in FM from 0.91 to 0.81 in the WDC dataset and from 0.87 to 0.69 in the Amazon-Google dataset. While the JSON format provides structured data, it does not inherently enhance LLMs' performance in ER tasks. The decrease in FM across datasets indicates that simpler, less structured formats like \emph{multi-attr} and \emph{multi-sim} are more effective for LLMs in ER.


\textbf{LLM-generated similarity scores create uncertainty: } The performance of \emph{multi-sim} shows that the similarity score generated from GPT-3.5 can create uncertainty in its results, even when selecting the optimal threshold value for maximizing FM. On the WDC dataset, the performance dropped from 0.91 to 0.71. However, the FM increased from the original method of 0.87 to 0.95 for the Amazon-Google Dataset. Therefore, this method shows variability but may be valuable in some contexts. Its cost was similar to \emph{multi-attr} (albeit the slight cost difference could become more magnified if the experiments are conducted on a large scale).

\textbf{The \emph{single-attr} prompt pattern is simple and viable prompt for performing cost-efficient ER if its underlying assumptions hold: } One of the goals behind this study was to determine the most cost-efficient method for ER, while controlling for relative gain in performance. We consider \emph{single-attr} and \emph{no-persona} as potential candidates, which are cheaper than other prompt patterns. We consider \emph{multi-attr} as a comparison-baseline due to its comprehensive attribute coverage, which tends to incur higher running costs. 
This setup allows us to directly assess the impact of attributes in the datasets and persona integration on ER efficiency, as \emph{multi-attr} can be simplified to \emph{single-attr} by removing all but the \textit{product title} attribute, and to \emph{no-persona} by omitting the \textit{persona} component. First, we compare the ER performance of \emph{multi-attr} with \emph{no-persona}, and found that removing persona negatively affects performance on a more difficult dataset such as Amazon-Google, where the FM decreased from 0.87 to 0.71. Additionally, the recall decreased to the lowest among all patterns, from 0.79 to 0.56. On the other hand, using \emph{single-attr} did not affect the performance as much. For the Amazon-Google dataset, the FM decreased from 0.87 to 0.81, while on the WDC dataset, \emph{single-attr} outperformed \emph{multi-attr}. This shows that \emph{single-attr}, despite its seeming simplicity, is a viable option for cost-efficient ER since no substantial difference is found (compared to \emph{multi-attr}) across the performance metrics; nevertheless, the cost when using a single attribute is considerably reduced (by about 37\%).We note, however, an important limitation here: \textit{single-attr} ultimately only works under the assumption that there is a single attribute that contains the information density for helping the LLM make the distinction between a match and a non-match; furthermore, it also assumes that this attribute is fixed across the dataset and that the domain expert has knowledge of it. In some situations, this assumption may prove restrictive and the advantages of \textit{multi-attr} may emerge more clearly.



\begin{table*}
\centering
\begin{tabular}{|p{0.70in}|p{0.7in}|p{0.75in}|p{0.7in}|p{0.7in}|p{0.7in}|p{0.7in}|}
    \hline
    \textbf{} & \textbf{multi-attr} & \textbf{single-attr} & \textbf{multi-json} & \textbf{few-shot} & \textbf{multi-sim}  \\
    \hline
    no-persona (A-G) & 0.000 & 0.000 & 0.019 & 0.000 & 0.000 \\
    \hline
    no-persona (WDC) & 0.000 & 0.000 & 0.068 & 0.000 & 0.000 \\
    \hline
\end{tabular}
\caption{A partial p-value matrix generated using paired Student's t-tests across the ER results between \textit{no-persona} versus all other five prompt patterns.}
\label{tab:p-value-matrix}
\end{table*}

\begin{table*}
\centering
\begin{tabular}
{@{}p{0.70in}@{}p{0.75in}@{}p{0.75in}@{}p{0.75in}@{}p{0.70in}@{}p{0.75in}@{}p{0.765in}@{}}
    \hline
    \textbf{} & \textbf{multi-attr} & \textbf{single-attr} & \textbf{multi-json} & \textbf{few-shot} & \textbf{multi-sim} & \textbf{no-persona} \\
    \hline
    multi-attr & 0.00/0.00 & 0.19/0.01 & 0.30/0.00 & 0.16/0.00 & 0.00/0.01 & 0.25/0.01 \\
    \hline
    single-attr & 0.10/0.01 & 0.00/0.00 & 0.27/0.00 & 0.15/0.00 & 0.01/0.01 & 0.26/0.00 \\
    \hline
    multi-json & 0.04/0.00 & 0.10/0.00 & 0.00/0.00 & 0.11/0.00 & 0.00/0.00 & 0.15/0.00 \\
    \hline
    few-shot & 0.17/0.00 & 0.24/0.00 & 0.37/0.00 & 0.00/0.00 & 0.01/0.00 & 0.33/0.00 \\
    \hline
    multi-sim & 0.20/0.01 & 0.29/0.01 & 0.46/0.00 & 0.21/0.00 & 0.00/0.00 & 0.00/0.01\\
    \hline
    no-persona & 0.02/0.01 & 0.12/0.00 & 0.18/0.00 & 0.10/0.00 & 0.00/0.01 & 0.00/0.00 \\
    \hline
\end{tabular}
\caption{Duplicate pairs / non-duplicate pairs in the ground-truth (Amazon-Google dataset) that the method in the row labeled correctly, but the method in the column labeled incorrectly, expressed as a fraction of the total number of duplicate and non-duplicate pairs in the ground-truth, respectively.}
\label{tab:confusion-matrix-ag}
\end{table*}

\begin{table*}
\centering
\begin{tabular}
{@{}p{0.70in}@{}p{0.75in}@{}p{0.75in}@{}p{0.75in}@{}p{0.70in}@{}p{0.75in}@{}p{0.765in}@{}}
    \hline
    \textbf{} & \textbf{multi-attr} & \textbf{single-attr} & \textbf{multi-json} & \textbf{few-shot} & \textbf{multi-sim} & \textbf{no-persona} \\
    \hline
    multi-attr & 0.00/0.00 & 0.02/0.02 & 0.22/0.01 & 0.07/0.02 & 0.04/0.03 & 0.07/0.01 \\
    \hline
    single-attr & 0.05/0.02 & 0.00/0.00 & 0.26/0.01 & 0.10/0.02 & 0.06/0.03 & 0.10/0.01 \\
    \hline
    multi-json & 0.01/0.01 & 0.02/0.01 & 0.00/0.00 & 0.04/0.01 & 0.02/0.01 & 0.03/0.01 \\
    \hline
    few-shot & 0.02/0.02 & 0.03/0.02 & 0.21/0.01 & 0.00/0.00 & 0.04/0.02 & 0.07/0.01 \\
    \hline
    multi-sim & 0.05/0.03 & 0.05/0.03 & 0.26/0.01 & 0.10/0.02 & 0.00/0.00 & 0.11/0.01\\
    \hline
    no-persona & 0.01/0.01 & 0.01/0.01 & 0.18/0.01 & 0.05/0.01 & 0.03/0.01 & 0.00/0.00 \\
    \hline
\end{tabular}
\caption{Duplicate pairs / non-duplicate pairs in the ground-truth (WDC dataset) that the method in the row labeled correctly, but the method in the column labeled incorrectly, expressed as a fraction of the total number of duplicate and non-duplicate pairs in the ground-truth, respectively.}
\label{tab:confusion-matrix-wdc}
\end{table*}


\textbf{Differences in ER results using \textit{no-persona} as a baseline: } Because many LLM prompting engineering papers rely on the persona as a means of improving the performance of the LLM, we also analyzed the \emph{no-persona} method by reporting the results of each of the five methods against the \emph{no-persona} method. Table \ref{tab:p-value-matrix} summarizes the p-value results for both datasets; for Amazon-Google, we found that the difference between \textit{no-persona} and four out of the five prompt patterns (except \emph{multi-json}) was significant at the 99.9\% confidence level or above. The \emph{multi-json} method was slightly less significant ($p= 0.019$). For WDC, the difference between \textit{no-persona} and four other prompt methods (except \emph{few-shot}), was significant at the 99.9\% confidence level or above. The \emph{few-shot} method was moderately significant against no-persona, with a p-value of 0.068. Even when the overall performance across prompting patterns is consistent, results from the paired t-test suggest that prompting patterns do affect the ER prediction results.

Given the statistical differences noted earlier, we report the analog of a `confusion matrix' in Table \ref{tab:confusion-matrix-ag} for the Amazon-Google dataset and Table \ref{tab:confusion-matrix-wdc} for the WDC dataset to better quantify disagreement between these methods. In both datasets, we found that the methods disagreed significantly less on the \emph{non-duplicates}, but that disagreement can be non-trivial and non-zero for duplicates. For instance, on duplicate pairs in the ground-truth that \emph{multi-sim} classified correctly, the \emph{multi-json} method (which used roughly the same information as \emph{multi-sim} in the prompt, but with more structure), classified 46\% of those pairs incorrectly (as non-duplicates) in Amazon-Google dataset and 26\% in WDC dataset. This suggests that consistency, while high overall, is not always guaranteed between different prompting methods. In the extreme case, we found that there were pairs where all methods agreed (and were incorrect), and cases where the methods were equally divided (obviously, with only some correct). Because each method also prompted the LLM for a reason for its output in its instructions, it is feasible to do a qualitative analysis in the near future of both (inter-prompt) disagreement, and of the LLM's rationale when it was consistently wrong, using the data released with this study.  

This confusion matrix supports the claim made earlier that \emph{single-attr} is a better option to perform cost-efficient ER compared to \emph{no-persona}. In the Amazon-Google dataset, as suggesyed by Table \ref{tab:confusion-matrix-ag}, \emph{single-attr} (on average) provides more correct ER results on pairs that other patterns get wrong, compared to \emph{no-persona}. Additionally, when comparing the row values of \emph{multi-attr} for the   \emph{single-attr} column (row 1, column 2) and \emph{no-persona} (row 1, column 6), the values are 0.19 and 0.25, respectively. This shows that out of all candidate pairs that \emph{multi-attr} provided correct decisions on, \emph{single-attr} made fewer mistakes when compared to \emph{no-persona}. We observe similar trends in Table \ref{tab:confusion-matrix-wdc} for the WDC dataset, suggesting the findings are stable and not just a measure of random fluctuation.

We end with a comment on why the costs listed in Table \ref{tab:prompt_eng_result} are not as trivial as they might seem. In practice, the records in the dataset would not be already paired (as was assumed for purposes of prompting the LLM) but would instead be represented simply as a set of $n$ records. For example, considering the Amazon-Google dataset, the Amazon partition has 1,363 records and the Google partition has 3,226 records. Without the blocking step (see Section 1), the total number of pairs to be evaluated would be more than 3 million (compared to fewer than 12,000 that we evaluated herein). As noted earlier, because the scope of this paper is limited to similarity, not blocking, we assumed perfect blocking and used a set of duplicate and non-duplicate pairs present in the ground-truth that the different prompting methods were evaluated on, similar to an ordinary binary classification problem. Even if blocking yielded 95\% reduction in the number of pairs, we would still need to prompt the LLM with almost 150,000 pairs, leading to 10-15x more expense than shown in Table \ref{tab:prompt_eng_result}, even for this relatively small dataset.


\section{Qualitative Analyses}

In the experimental results described earlier, we gained insights into the overall ER performance and cost of each prompt pattern through quantitative and statistical analyses. In this section, we conduct selective \textit{qualitative} analysis of specific candidate pairs to understand the differences and similarities between the prompt patterns, as well as draw some broader conclusions about GPT-3.5 by focusing on both success and failure cases. 

\begin{figure*}[ht]
    \centering\footnotesize
    \includegraphics[width=0.95\textwidth]{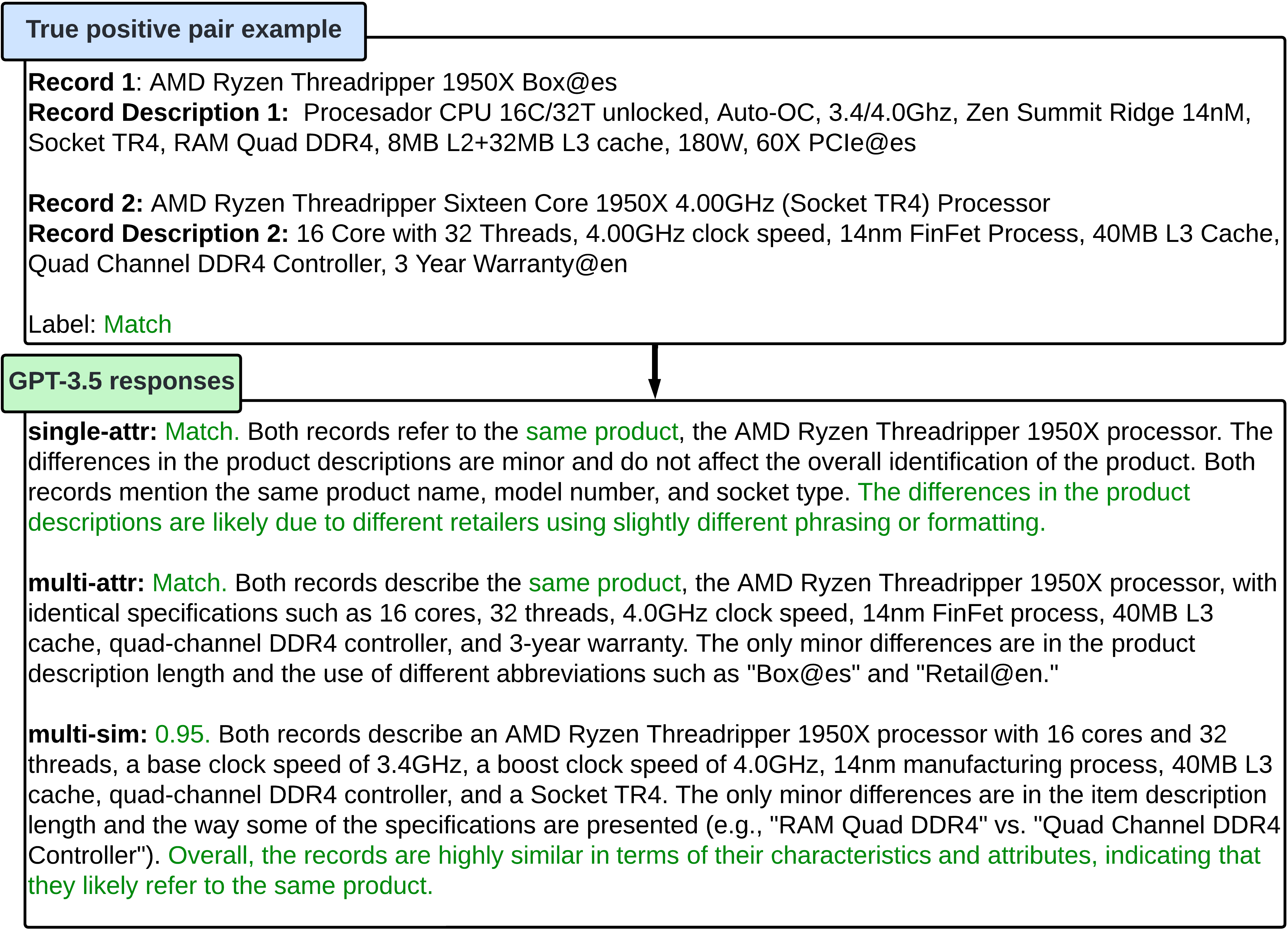}
    \caption{An illustration of a relatively `easy' duplicate pair, where the two records differ in naming and description order. GPT-3.5 accurately concludes this pair as a duplicate using all prompt patterns, with the correct ER results and explanations highlighted in green.} 
    \label{fig:TP}
\end{figure*}

We begin by discussing some of GPT-3.5's success cases on the ER benchmarks, by considering the candidate pairs where GPT-3.5 successfully outputs true positives (TP) and true negatives (TN) for \textit{all} six prompt patterns. Figure \ref{fig:TP} shows a candidate pair consisting of (1) AMD Ryzen Threadripper 1950X Box@es and (2) AMD Ryzen Threadripper Sixteen Core 1950X 4.00GHz (Socket TR4) Processor, where both entries refer to the same real-world product but with slight differences in naming. Arguably, this is an `easy' case, and we find (expectedly) that the \textit{single-attr} response demonstrates GPT-3.5's ability to handle such pairs, showing that slight variations in phrasing and formatting do not negatively affect the ER result. Similar responses are observed using the \textit{multi-attr} and \textit{multi-sim} patterns, in which the pairs include additional details from the description, manufacturer, and price attributes, yet GPT-3.5 still accurately decides the correct ER result. All other prompt methods showed the same outcome and explanation, so we omitted the responses due to space limitations. This result shows that GPT-3.5 does not get confused easily on pairs that are relatively obvious (even when presented with extraneous information), which marks an advance from methods like string matching, which would presumably get more confused due to inclusion of irrelevant attributes and information, at least for the purposes of determining whether the pairs are matching or not.

\begin{figure*}[ht]
    \centering\footnotesize
    \includegraphics[width=0.95\textwidth]{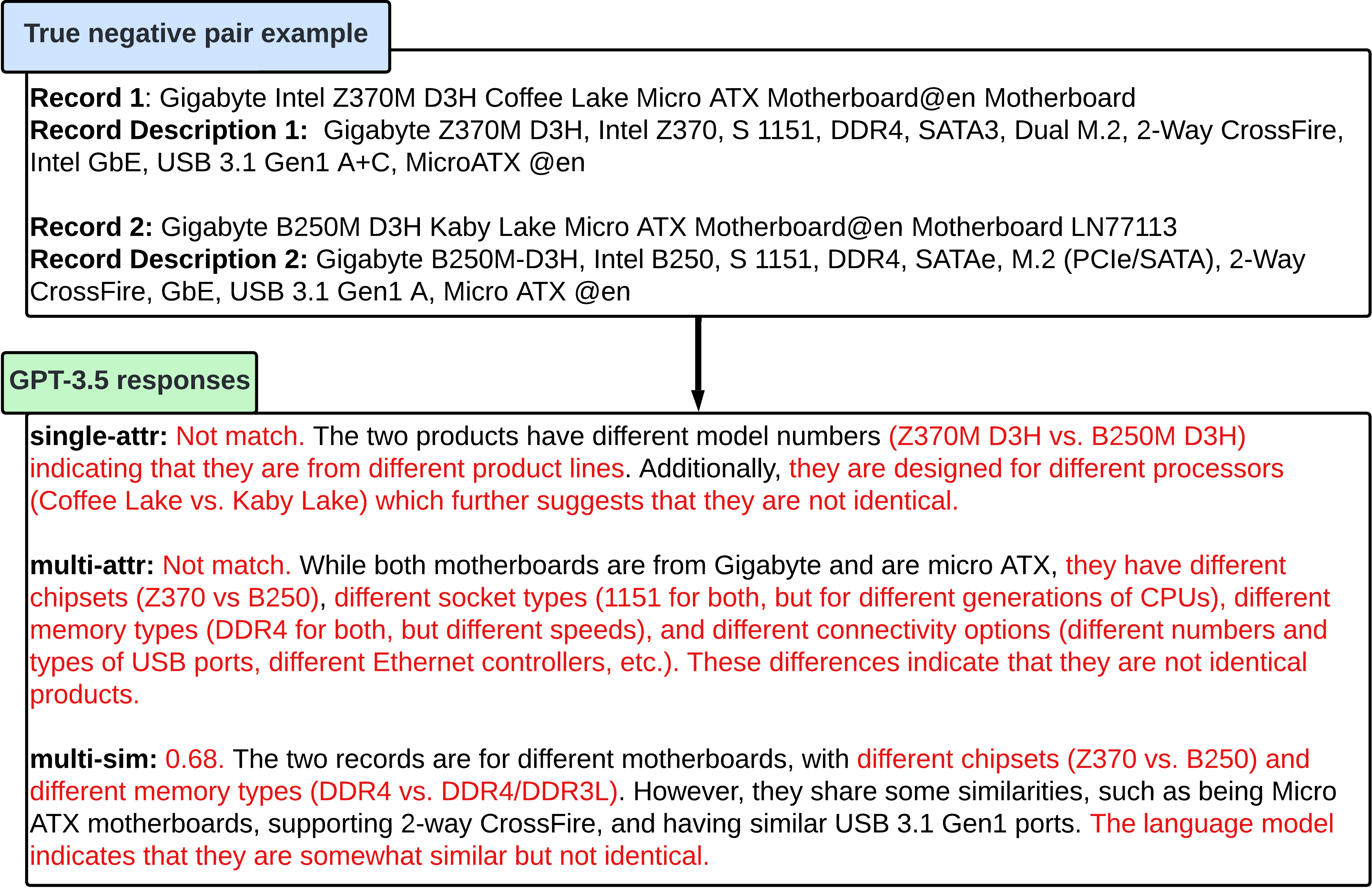}
    \caption{An illustration of a relatively `easy' non-duplicate pair, where the two records clearly differ in their model numbers. GPT-3.5 accurately concludes this pair as a non-duplicate using all prompt patterns, with the correct ER results and explanation indicated with red highlights.} 
    \label{fig:TN}
\end{figure*}

Similar observations hold for `easy' non-duplicate pairs. Figure \ref{fig:TN} shows such a pair consisting of (1) Gigabyte Intel Z370M D3H Coffee Lake Micro ATX Mother-board@en Motherboard and (2) Gigabyte B250M D3H Kaby Lake Micro ATX Mother-board@en Motherboard LN77113 where a clear distinction is in their product number (Z370M vs B250M). The \textit{single-attr} prompt pattern is again found to be able to differentiate between these products based on their titles, pinpointing the chipset variants Z370M and B250M and elaborating that \textit{Coffee Lake} and \textit{Kaby Lake} refer to distinct processor designs. Furthermore, responses using the \textit{multi-attr} and \textit{multi-sim} prompt patterns demonstrate GPT-3.5's capability to incorporate record descriptions into the decision-making and further contextualize the explanation by comparing additional features like memory types, connectivity options, and USB ports between the two motherboards.

\begin{figure*}[ht]
    \centering\footnotesize
    \includegraphics[width=0.95\textwidth]{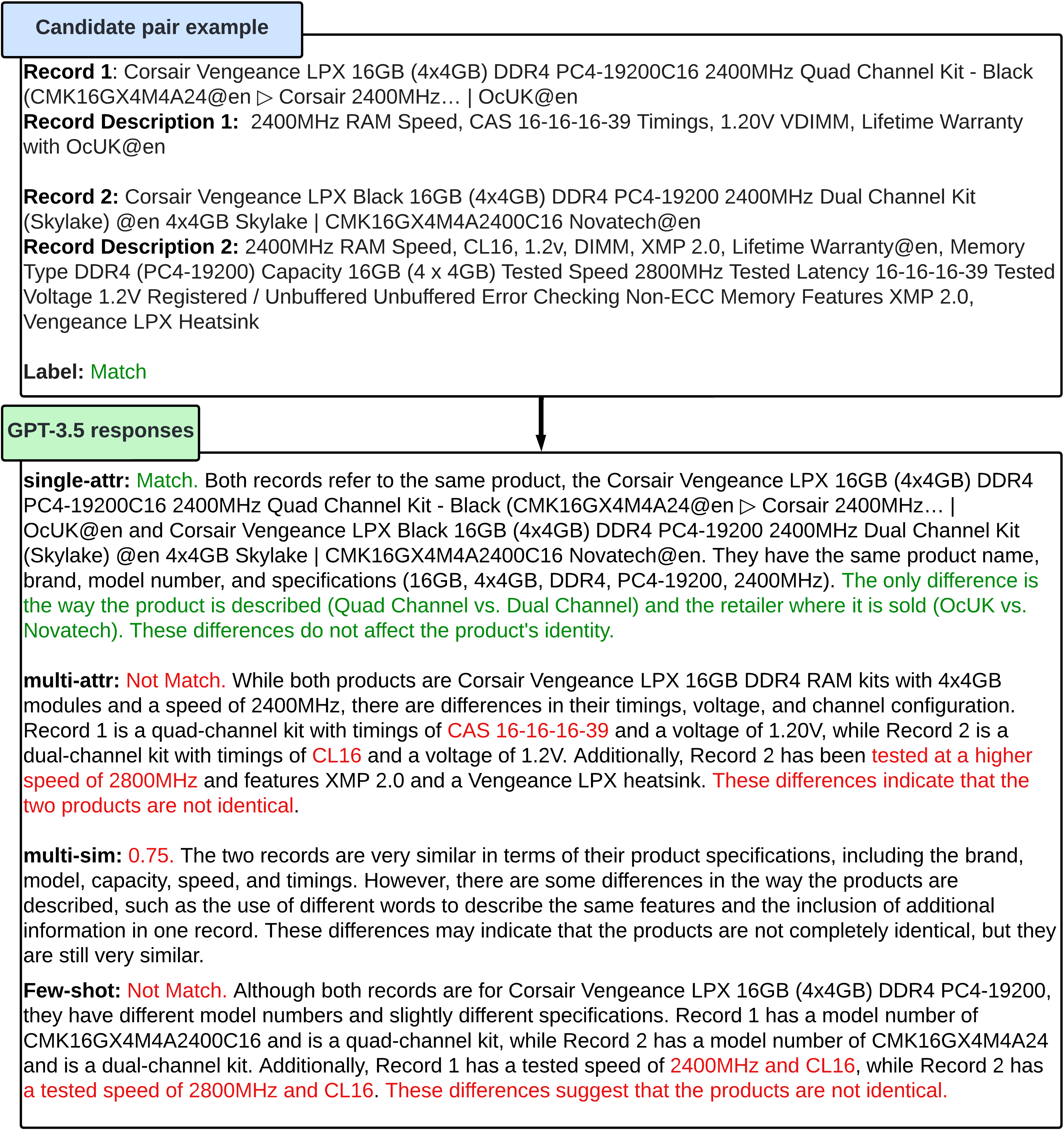}
    \caption{An illustrative example of a duplicate pair with disagreement on ER results from different prompt patterns. This candidate pair is more `difficult', containing complex technical terminology in the title and description. Of the six patterns, only one correctly identifies the pair as duplicates, showing the challenges GPT-3.5 faces when processing domain-specific lingo. We highlight the wrong explanations from GPT-3.5 in red.} 
    \label{fig:c-2}
\end{figure*}

Next, we discuss cases and examples of candidate pairs that GPT-3.5 found more challenging. As shown in the example in Figure \ref{fig:c-2}, GPT-3.5 may provide inaccurate responses when processing pairs that include complex technical terminology or jargon. The figure shows a duplicate pair of (1) Corsair Vengeance LPX 16GB 2400MHz Quad Channel Kit and (2) Corsair Vengeance LPX 16GB 2400MHz Dual Channel Kit, with details such as full names and descriptions provided in the figure. These two products have differences in their channel kit description (dual vs quad channel), which explains the number of RAM channels tested to operate simultaneously in the product. However, this information does not differentiate the product identity. The response from \textit{single-attr} concludes the correct result and explains that the small difference in dual channel vs quad channel does not affect the identity of the product. However, all other prompt patterns with multiple attributes made incorrect decisions. This is due to two factors: firstly, the description of record 2 contains a tested speed of 2800 MHz that does not exist in the record 1 description. This confused GPT-3.5 with the actual speed that both products have at 2400MHz. Secondly, the difference in CAS timing format between the two records causes GPT-3.5 to misinterpret CL16 as different from 16-16-16-19 CAS timing. These observations highlight how discrepancies in technical specifications and formatting can lead GPT-3.5 to misinterpret critical details, resulting in incorrect ER decisions.

\begin{figure*}[ht]
    \centering\footnotesize
    \includegraphics[width=0.95\textwidth]{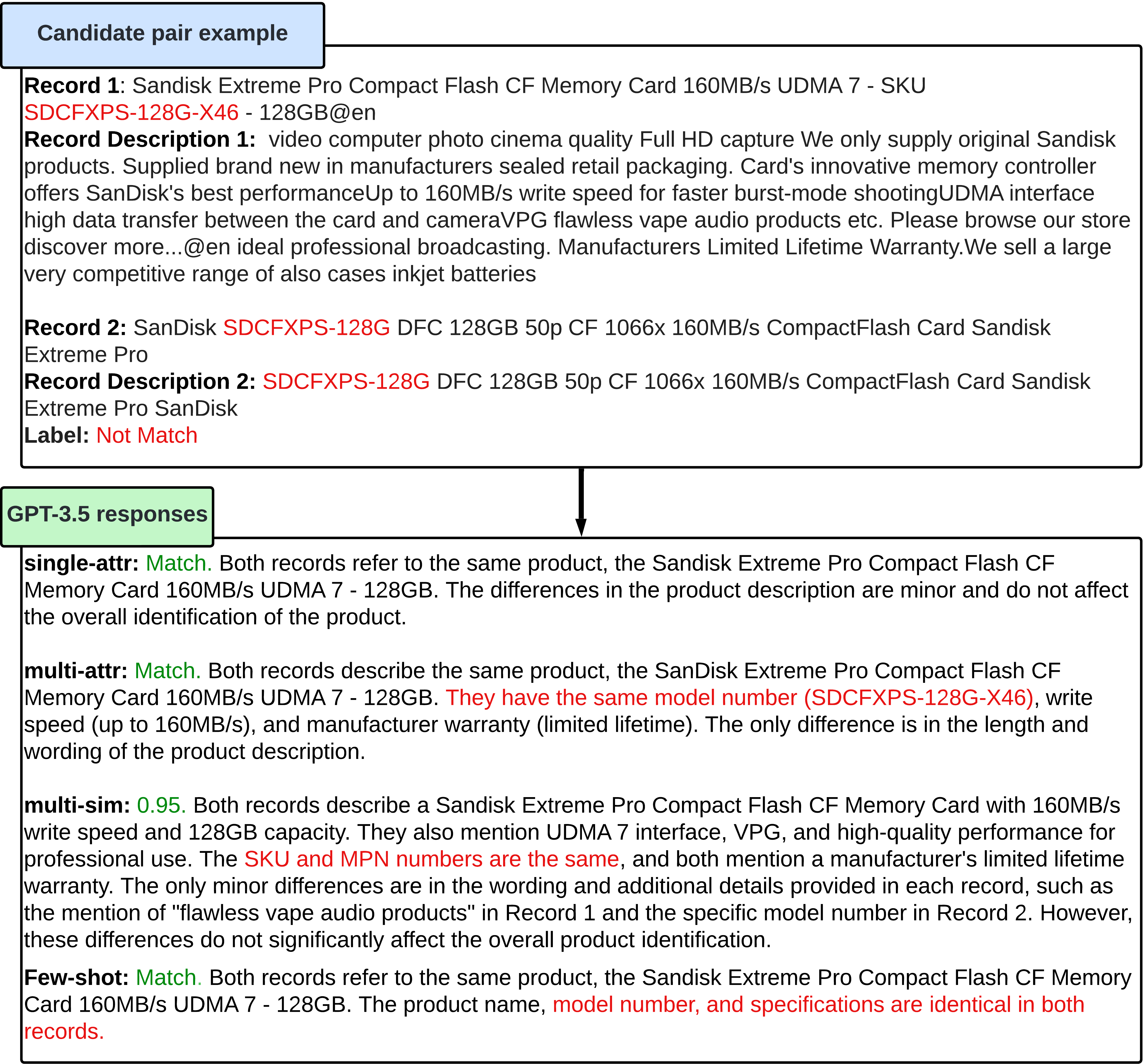}
    \caption{An illustrative example of a non-duplicate pair with disagreement on ER results from different prompt patterns. This pair contains a minor difference in model numbers, causing GPT-3.5 to make mistakes. The responses across all prompt patterns hallucinated and provided incorrect explanations (highlighted in red), pointing out (incorrectly) that the two products have the same model numbers.} 
    \label{fig:c-3}
\end{figure*}

Another challenging candidate pair is shown in Figure \ref{fig:c-3}, representing a non-duplicate pair, where the only difference between the two records is the model number (SDCFXPS-128GB-X46 vs SDCFXPS-128GB). This pair presents a candidate pair that is also challenging for humans, who typically do not memorize details such as model numbers. GPT-3.5 also struggled to provide the correct result, where all prompt patterns provided incorrect ER results. The explanations from all patterns hallucinated and mentioned that the model numbers of the two products are the same. This shows that GPT-3.5 still lacks the knowledge to differentiate products at the granular level of model numbers. 

In considering both qualitative and quantitative results, therefore, we find an increase in candidate pair attributes doesn't necessarily translate to better performance for GPT-3.5. The \emph{multi-json} prompt pattern even shows a decrease in performance, likely due to the complexity and structure of JSON affecting GPT-3.5's processing capabilities. The LLM's ability to generate similarity scores (e.g., when using \emph{multi-sim}) demonstrates that it can achieve competitive ER performance, even in the absence of predefined similarity metrics. However, it introduced a level of uncertainty, suggesting that GPT-3.5-generated approximations can be ambiguous and not always reliable. The \textit{single-attr} prompt pattern, with its simplicity and relative cost-efficiency, emerged as a viable and cost-efficient option for ER under the assumption that we know a single attribute that contains the most information density for enabling the LLM to make ER decisions. 

Qualitatively, GPT-3.5 demonstrated consistent performance in processing pairs that are either clearly identical or distinct, indicating robustness in handling straightforward ER tasks. Although we provided only two detailed examples of such pairs, such results are found to be common across the full dataset. We refer readers interested in the comprehensive results to the supplementary materials included with this paper. In contrast, the model struggled with intricate details such as product model numbers, testing methods, and specifications. This suggests a gap in the LLM's deep knowledge or understanding when processing pairs that are highly correlated but have slight differences in their detail. Furthermore,  excessive information can sometimes leads to confusion, indicating that an overload of data can negatively direct GPT-3.5 to make incorrect ER decisions. These insights collectively highlight the nuanced balance between the quantity of information provided and the quality of ER outcomes, and may point toward more optimized prompting strategies for LLMs in future ER applications.

\section{Conclusion}\label{sec:conclusion}

This experimental study sought to evaluate whether a recent LLM like GPT-3.5 could offer a promising, cost-efficient and unsupervised alternative to conventional ER similarity functions, which generally require training data (and computation time) and domain-specific feature engineering. In addition to comparing the costs and performance of these methods, we also quantified the extent to which these models disagreed. Our results showed that, while the prompting method does matter, the results are generally stable and consistent (but that disagreement can be high between some prompting methods, particularly on duplicates). Furthermore, there can be significant variance in cost, which becomes important at the scales for which industrial ER is typically designed.  While integrating blocking into an LLM-based ER problem remains an open problem at present, we hope that this study can provide guidance on efficiently using LLMs for unsupervised ER similarity.  To support further analysis and replication, we have collected and will release all primary data underlying our analyses as supplementary materials.

\section*{Data Availability}
All raw data and analyses cited and used in the main text may be accessed here: \url{https://drive.google.com/drive/folders/18taqVQ8oJeNunMb6nz_qZNZ1EnYy7JCF?fbclid=IwAR0jLxyoZQgUbI7wtcL-mbIM6YWdH0L8QYeg99ZKl2xP3FiA4pmrt4Op99Q}.

\bibliography{pr-bib}
\bibliographystyle{plain}

\end{document}